\pdfoutput=1

\documentclass[11pt]{article}

\usepackage[final]{acl}

\usepackage{times}
\usepackage{tabularx} 
\usepackage{latexsym}
\usepackage{booktabs} 
\usepackage{array}     
\usepackage[T1]{fontenc}

\usepackage[utf8]{inputenc}
\usepackage{pifont}
\usepackage{amssymb}

\usepackage{microtype}
\usepackage{hyperref}
\usepackage{inconsolata}

\usepackage{graphicx}
\usepackage{caption}

\title{Deriving Strategic Market Insights with Large Language Models: \\ A Benchmark for Forward Counterfactual Generation}

\author{
    Keane Ong\textsuperscript{$\spadesuit$}\textsuperscript{$\diamondsuit$}, 
    Rui Mao\textsuperscript{$\clubsuit$}, 
    Deeksha Varshney\textsuperscript{$\spadesuit$}\textsuperscript{$\heartsuit$},
    Paul Pu Liang\textsuperscript{$\diamondsuit$},
    \\ \bf Erik Cambria\textsuperscript{$\clubsuit$} and 
    Gianmarco Mengaldo\textsuperscript{$\spadesuit$}\thanks{Corresponding author: mpegim@nus.edu.sg} \\
    \textsuperscript{$\spadesuit$}National University of Singapore
    \textsuperscript{$\diamondsuit$}Massachusetts Institute of Technology \\
    \textsuperscript{$\clubsuit$}Nanyang Technological University    \textsuperscript{$\heartsuit$}Indian  Institute of Technology, Jodhpur \\
    \texttt{keane.ongweiyang@u.nus.edu};
     \texttt{deeksha@iitj.ac.in}; \texttt{\{mpegim\}@nus.edu.sg};\\
     \texttt{ppliang@mit.edu};
     \texttt{\{rui.mao,cambria\}@ntu.edu.sg}
}

\begin{document}
\maketitle
\begin{abstract}
Counterfactual reasoning typically involves considering alternatives to actual events. While often applied to understand past events, a distinct form--forward counterfactual reasoning--focuses on anticipating plausible future developments. This type of reasoning is invaluable in dynamic financial markets, where anticipating market developments can powerfully unveil potential risks and opportunities for stakeholders, guiding their decision-making. However, performing this at scale is challenging due to the cognitive demands involved, underscoring the need for automated solutions. LLMs offer promise, but remain unexplored for this application. To address this gap, we introduce a novel benchmark, \textsc{Fin-Force}--\textbf{FIN}ancial \textbf{FOR}ward \textbf{C}ounterfactual \textbf{E}valuation. By curating financial news headlines and providing structured evaluation, \textsc{Fin-Force} supports LLM based forward counterfactual generation. This paves the way for scalable and automated solutions for exploring and anticipating future market developments, thereby providing structured insights for decision-making. Through experiments on \textsc{Fin-Force}, we evaluate state-of-the-art LLMs and counterfactual generation methods, analyzing their limitations and proposing insights for future research. We release the benchmark, supplementary data and all experimental codes at the following link:~\url{https://github.com/keanepotato/fin_force}
\end{abstract}

\vskip 0.1cm

\section{Introduction}
Counterfactual reasoning--considering alternatives to actual events--allows us to envision possibilities beyond reality~\cite{byrne2016counterfactualdef}. While often used retrospectively to consider \textit{what could have happened}, a distinct forward-looking form--\textit{forward counterfactual reasoning}--focuses on \textit{what could happen next}~\cite{todorova2015counterfactualfuture, bynum2023counterfactualsfuture}. 

\begin{figure}[h]
    \centering
    \includegraphics[width=\linewidth]{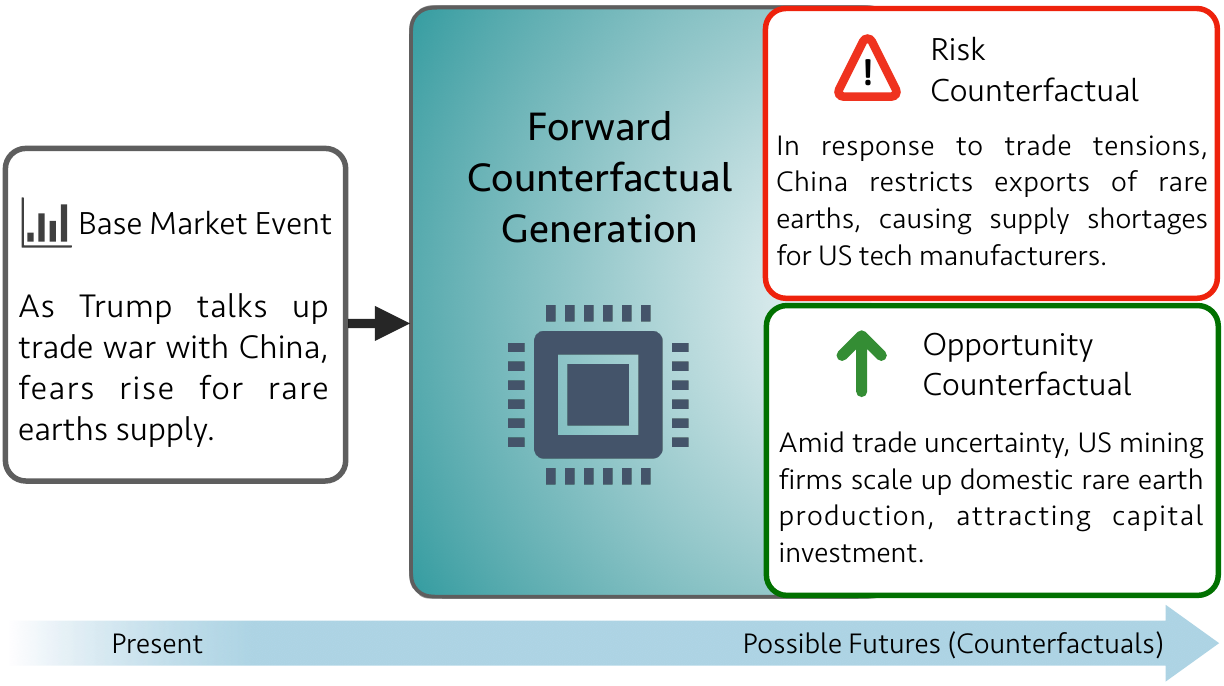}
    \caption{Overview of \textsc{Fin-Force} task. Given a financial news headline depicting a market event, an LLM is tasked with generating two forward counterfactuals - an opportunity counterfactual and a risk counterfactual. While the opportunity counterfactual explores how the event can positively shift, the risk counterfactual highlights potential adverse scenarios.}
    \label{fig:overview}
\end{figure}

In dynamic financial markets, forward counterfactual reasoning often plays a crucial role in guiding decision-making~\cite{byrne2016counterfactualdef,du2024financialsentiment}. 
To this end, market stakeholders frequently anticipate future developments from current events--not to predict exact outcomes, but to explore plausible future scenarios that inform strategic responses. This allows them to hedge against potential risks and capitalize on emerging opportunities~\cite{greenwood2014financeexpectations}. 
Yet, despite its strategic value, forward counterfactual reasoning remains difficult to scale. Specifically, due to its complex cognitive demands--i.e., requiring reasoning across multiple causal relationships~\cite{lebow2000counterfactualdifficult}--it is impractical to apply forward counterfactual reasoning on an extensive range of market events or within short timeframes. 
In other words, stakeholders face constraints in how widely and rapidly they can explore and anticipate future developments from current events. This limits their ability to derive timely foresight for strategic decision-making, leaving them exposed to missed opportunities and strategic missteps~\cite{schoemaker1995scenarioplanning}.

Addressing this need for scalability, automated tools--such as LLMs--can support forward counterfactual reasoning. Accordingly, LLMs can enable the expansive generation of plausible future developments based on actual events~\cite{duuret}. 
However, LLM counterfactual research has often focused on specific applications, such as narrative rewriting or text classification~\cite{wang2024surveycounterfact}. The use of LLMs to support forward counterfactual reasoning in finance, or any domain, remains largely unexplored.
To bridge this gap, we propose \textsc{Fin-Force} -- \textbf{FIN}ancial \textbf{FOR}ward \textbf{C}ounterfactual \textbf{E}valuation, a novel benchmark to support LLM forward counterfactual generation in finance. 

\textsc{Fin-Force} comprises news headlines, each describing a distinct market event, which serves as a basis for LLMs to generate two types of forward counterfactuals. i) The \textit{opportunity counterfactual} explores how the market event could positively shift, resulting in favorable implications for market stakeholders.
ii) The \textit{risk counterfactual} highlights how the market event could adversely shift, exposing vulnerabilities and negative market implications. While opportunity counterfactuals anticipate potential upside scenarios for stakeholders to capitalize on, risk counterfactuals identify emerging risks for stakeholders to hedge against (Figure~\ref{fig:overview}). By laying the foundation for LLMs to generate these forward counterfactuals, \textsc{Fin-Force} supports scalable and automated insights into potential market opportunities and risks before they materialize, enhancing stakeholders' strategic decision-making.

Through experiments, we evaluate a wide range of LLM-based methods on the \textsc{Fin-Force} benchmark to provide insights for future model development. Our findings highlight: (1) LLMs, under zero-shot and few-shot prompting, do not perform equally well on \textsc{Fin-Force}, with Claude 3.5 Haiku performing better than Qwen 2.5 72B, Llama 4-Maverick, Gemini 2.0 Flash and GPT-4o. (2) The evaluated state-of-the-art (SOTA) counterfactual prompting methods perform poorly, while SOTA sampling-based counterfactual generation achieves the best performance. (3) A self-training paradigm can enable a smaller LLM (i.e. Llama3.1 8B) to outperform all large-scale LLMs (i.e. GPT-4o) under zero-shot and few-shot prompting. (4) The limitations of the different methods via qualitative analysis and sub-task performance, and research directions for tackling these limitations.

$ $

We summarize our main contributions as follows. (1) We develop and release \textsc{Fin-Force}, a novel benchmark comprising 1368 news headlines that describe market events, to support forward counterfactual generation in finance. (2) We conduct extensive experiments on \textsc{Fin-Force} to evaluate a wide-range of methods, offering insights to guide future model development in the NLP community. While both contributions are situated in the financial domain, the underlying task of forward counterfactual generation is, to our knowledge, one of the first task formulations of its kind. The task and its core principles--projecting plausible futures, and directional outcomes (positive and negative)--could be potentially extended to other complex and dynamic domains such as public policy~\cite{tetlock2017expert} or scenario planning~\cite{schoemaker1995scenarioplanning}, to support strategic decision-making.

$ $

\section{Related Work}
\textbf{Counterfactual Reasoning.} Counterfactual reasoning involves considering alternative outcomes based on changes to a given “base” situation~\cite{byrne2016counterfactualdef}. It is used to explain past events, anticipate possible future developments, and support a variety of tasks across domains~\cite{byrne2016counterfactualdef,wang2024surveycounterfact}. In decision-making contexts, anticipating future developments is particularly valuable in uncertain environments beyond finance, including policy~\cite{tetlock2017expert} and scenario planning~\cite{schoemaker1995scenarioplanning}. While our work centers on the financial domain, the task we develop--scalable counterfactual generation for projecting plausible future developments (i.e. forward counterfactual generation)--could be extended to support strategic decision-making in other complex, uncertain settings.

\noindent \textbf{NLP Counterfactual Generation Benchmarks.}
Counterfactual generation has been widely studied in NLP, with established benchmarks supporting different applications. \textsc{TimeTravel} focuses on counterfactual story rewriting~\cite{qin2019timetravelcf}, \textsc{SNLI} explores modifications to premises or hypotheses for natural language inference~\cite{kaushik2019snli}, and \textsc{CounterFact} evaluates whether language models can faithfully update specific factual knowledge~\cite{meng2022counterfact}. Despite the progress, existing benchmarks do not consider temporal progression from a base event or forward-looking counterfactuals that project how events might plausibly unfold from present scenarios. Additionally, leveraging counterfactual generation for practical financial applications remains largely unexplored. To address these gaps, we introduce \textsc{Fin-Force}, a novel benchmark for forward counterfactual generation in finance.

\noindent \textbf{LLM Methods for Counterfactual Generation.} 
LLMs’ strength in counterfactual generation has spurred methods such as prompt engineering~\cite{nguyen-etal-2024-llms_for_cfs}, mask-and-replace~\cite{feng-etal-2024-counterfactualdistllation}, anomalous language modeling~\cite{mao2024metapro2}, and controlled text generation~\cite{ravfogelgumbel}. The generalizability of these methods to new tasks remains unexplored, and state-of-the-art LLM paradigms like self-training~\cite{SRLM} have yet to be applied to counterfactual generation. Our work extends the literature by evaluating the generalizability of existing LLM counterfactual methods to forward counterfactual generation. We also explore the potential and limitations of self-training when applied to a counterfactual generation setting.

\noindent \textbf{LLM Applications in Financial Tasks.} Recent advances in LLMs have enabled a range of promising applications in finance~\cite{du2025finnlp,yeo2023comprehensive}, thanks to LLMs’ strong reasoning abilities~\cite{plaat2024llmreasoning}. These include stock prediction~\cite{heng2025llmstockpred}, financial statement analysis~\cite{kim2024financial}, ESG rating evaluation~\cite{ong2025xnlp}, and greenwashing detection~\cite{ong2025greenwash}. However, the counterfactual reasoning capabilities of LLMs—and specifically their capacity to perform scenario analysis (i.e., exploring possible future risks and opportunities)—remain underexplored. This is despite the central role that scenario analysis plays in the financial sector, including within investment funds that proactively anticipate future market trends based on evolving narratives~\cite{phadnis2015effectscenario}. By investigating LLMs’ abilities in this complex area, we extend research on LLM-based financial applications to advanced reasoning tasks.

\section{Benchmark Construction}\label{sec:benchmark_construct}
The construction of \textsc{Fin-Force} begins with a broad collection of news data, prior to annotating news that describe market events. These annotated news are then included in the benchmark.

\subsection{News Collection}\label{sec:news_collection}

To collect news headlines that describe financial market events, we queried {NewsAPI}\footnote{\url{https://newsapi.org}} with market-related keywords (eg., \textit{GDP growth}, \textit{Interest rates}--full-list in Appendix Table~\ref{tab:market_keywords}). These keywords are chosen to capture information generally associated with the financial markets. The duration from which we collected news headlines spans from 1 September 2024 to 18 April 2025. This period was selected to post-date the knowledge cut-offs or release dates of the LLMs evaluated in Section~\ref{sec:expts}, ensuring no overlap with their pretraining data. This improves the robustness of our findings.

\subsection{News Annotation Scheme}
Given that not all headlines collected from NewsAPI pertain to financial market events, we apply human annotation to select those that clearly do. Accordingly, a headline is selected for the \textsc{Fin-Force} benchmark if it meets the following criteria\footnote{Details of the annotators, annotation instructions, descriptions of the market categories are in Appendix~\ref{appendix:benchmark}.}, as summarized below:

\noindent \textbf{Event Status:}
The headline must explicitly describe an ongoing financial market event. An event is defined as a current, factual development representing a specific change or occurrence affecting financial markets or the broader economy. Examples include central bank decisions, regulatory changes, or macroeconomic data releases. Headlines reflecting generic commentary or opinions without concrete developments are excluded.

\noindent \textbf{Material Relevance:}
The event must be material to market participants, meaning it reflects a development that can potentially influence financial decision-making or market outcomes. This includes broad systemic drivers (e.g., monetary policy changes, macroeconomic indicators) and significant corporate events (e.g., mergers, earnings surprises, corporate restructuring). These event types are described by the market categories in Table~\ref{tab:stats_market_categories}. To be considered material, the event must explicitly relate to at least one of these market categories\footnotemark[\value{footnote}].


\begin{table}[h!]
\centering
\begin{tabularx}{\linewidth}{@{}Xr@{}}
\toprule
\textbf{Market Categories} & \textbf{Count} \\
\midrule
Monetary policy \& central banking & 109 \\
Corporate strategy \& operations & 229 \\
Geopolitical \& regulatory developments & 148 \\
Financial markets \& asset performance & 574 \\
Supply chain \& logistics & 24 \\
ESG \& sustainability developments & 13 \\
Technology \& innovation & 22 \\
Labour \& employment & 27 \\
Macroeconomic indicators & 186 \\
Banking \& financial stability & 36 \\
\bottomrule
\end{tabularx}
\caption{\textsc{Fin-Force} categories and headline counts. Certain topics appear more frequently due to their natural prevalence in financial news, resulting in a benchmark that better reflects real-world news distributions.}\label{tab:stats_market_categories}
\end{table}

\subsection{News Annotation Process}\label{sec:annotation_process}
Our annotation process\footnotemark[\value{footnote}] involves 3 annotators and 2 verifiers, all of whom are doctoral or post-doctoral researchers specializing in finance. (1) Training Phase: Annotators and verifiers participate in iterative trial rounds, each consisting of 60 randomly selected samples. 

After each round, annotations are reviewed and feedback given, continuing until participants reach $\ge$95\% accuracy.
(2) Annotation: Once the trial phase is complete, annotators proceed with daily labeling. They are instructed to flag any samples where they are uncertain about the correct label.
(3) Disagreement Resolution: Every two days, flagged samples are reviewed in group discussions among the annotators to reach a consensus. If unanimity is not possible, the majority decision is adopted.
(4) Verification: Every two days, 25\% of the annotated samples from each annotator (excluding those marked as uncertain) are randomly selected and reviewed by the verifiers for correctness and adherence to the guidelines. If more than 5\% of the reviewed annotations are found to be incorrect, the entire batch from that two day window is re-annotated. Our final benchmark comprises 1368 news headlines that describe financial market events, with a full breakdown shown in Table~\ref{tab:stats_market_categories}. 

\subsection{Supplementary Dataset}\label{sec:supplementary_set}
To supplement the benchmark, we annotate an additional 2,105 news headlines from 1 Jan 2021 to 20 June 2024, following the same annotation procedures. We use GPT-4o to generate forward counterfactuals for these headlines, creating a supplementary synthetic dataset\footnote{Supplementary dataset is also released at \url{https://github.com/keanepotato/fin_force} to support self-training model development. Further details of this dataset are in Appendix~\ref{appendix:supplementary_dataset}} for the SFT warm-up phase in our self-training paradigm (Section~\ref{sec:expts}). 

\subsection{Metrics Design}\label{sec:metrics}

Counterfactuals are traditionally evaluated using \textit{validity}, \textit{similarity}, \textit{diversity}, and \textit{fluency}~\cite{wang2024surveycounterfact}. However, not all are relevant to \textsc{Fin-Force}, which involves generating risk and opportunity counterfactuals from a market event described by a news headline. \textit{Validity}--whether the counterfactual flips a classifier’s label~\cite{mothilal2020explainingcounterfactuals}--is irrelevant, as our task does not involve predefined classifier labels that can be altered. \textit{Similarity}--the minimality of counterfactual edits from the original text~\cite{treviso2023crest}--can be counterproductive, as the blanket penalization of edits may discourage changes that introduce meaningful risk or opportunity shifts. \textit{Diversity}--divergence in semantic meaning between counterfactuals~\cite{wang2024surveycounterfact}--overlooks how the counterfactuals must represent distinct market risks and opportunities rather than \textit{any} difference in wording or semantics. \textit{Fluency}--naturalness and grammaticality of counterfactuals, via perplexity~\cite{radford2019perplex}--is the only metric which remains relevant.

Therefore, for our task, we build on perplexity for fluency evaluation by focusing on $\Delta$ Perplexity\footnote{Perplexity is computed by GPT-2; the average perplexity of all the original headlines is 267.98, averaged over ten runs.}--the difference between the average perplexity of all counterfactuals and all original headlines--which normalizes for the fluency of the original headlines. In place of \textit{validity}, \textit{similarity}, and \textit{diversity}, we introduce two new metrics--\textit{forward-compatibility} and \textit{directionality}, which are better aligned with FIN-FORCE's objectives. 
Importantly, these metrics assess whether counterfactuals reflect plausible future developments, not whether they actually occur. This is consistent with real-world strategic analysis, where even future developments that do not materialize can powerfully inform decision-making -- provided that they reflect credible future scenarios~\cite{schoemaker1995scenarioplanning}.

\textit{Forward-Compatibility}\footnote{Prompts for LLM evaluation, details about the human validation study are in Appendix~\ref{appendix:benchmark} \&~\ref{appendix:expts}. Further details on the new metrics can also be found in the human study.} assesses whether the counterfactual represents a plausible future development that logically follows from the original market event. It must meet: (i) Consistent progression. It reflects a continuation logically connected the original market event. (ii) Non-contradiction. It does not negate the original event by introducing mutually exclusive scenarios.

\textit{Directionality}\footnotemark[\value{footnote}] assesses whether the counterfactual shows a clear and meaningful market shift--toward either improvement (opportunity) or deterioration (risk) in market conditions--relative to the original market event. It must meet:
(i) Relative Significance. The shift is substantial compared to the original market event. (ii) Logical Soundness. The shift is grounded in logical reasoning that does not reflect economic or financial implausibility.
(iii) Scope of Impact. The shift extends beyond isolated parties to affect other market participants. (iv) Financial Consequence. The shift entails financial effects likely to influence market behaviour or decision-making.

Each counterfactual is assessed using an LLM-as-a-judge framework\footnotemark[\value{footnote}] with GPT-4o~\cite{zheng2023judgingllm}, and is labeled as satisfying \textit{forward-compatibility} or \textit{directionality} only if it meets all their corresponding criteria. The scores in our results--Fwd-Compat. and Dir. (Section~\ref{sec:results})--reflect the proportion (in percentages) of counterfactuals meeting the \textit{forward-compatibility} and \textit{directionality} criteria respectively. While LLM evaluation increases impartiality and scalability~\cite{zheng2023judgingllm}, we acknowledge its limitations. To ensure robustness, we validate the LLM judgments through a human-LLM agreement study\footnotemark[\value{footnote}] with 500 random samples. This study compares LLM labels with those from independent human annotators, showing an average agreement of 81.4\% on \textit{directionality} and 89.6\% on \textit{forward-compatibility}.

$ $

\section{Experiments}\label{sec:expts}
The following models are tasked with generating a single risk and a single opportunity counterfactual from each headline in \textsc{Fin-Force}. 

\noindent \textbf{Baseline LLM Prompting\footnote{Full details on how we adapted the methods for our task are in Appendix~\ref{appendix:model}.}.} Latest LLMs are evaluated under zero and few-shot prompting. To improve experimental robustness, selected models have release or knowledge cut-off dates preceding 1 September 2024 (\textsc{Fin-Force}'s headlines are collected after this date). LLMs include proprietary--Claude 3.5 Haiku~\cite{claude-3.5-haiku}, Gemini 2.0 Flash~\cite{gemini-2.0-flash}, GPT-4o~\cite{gpt-4o}, and open-source models--Llama 4 Maverick~\cite{Llama-4}, Qwen 2.5 72B~\cite{yang2024qwen2.5}.

\noindent \textbf{SOTA Counterfactual Generation\footnotemark[\value{footnote}].} We evaluate SOTA counterfactual generation algorithms adaptable to our task. LLMs-for-CFs~\cite{nguyen-etal-2024-llms_for_cfs} uses chain-of-thought prompting to identify and replace keywords in the original text. CounterfactualDistil~\cite{feng-etal-2024-counterfactualdistllation} masks topic words and noun phrases in the original text, then prompts an LLM to generate replacements. LM-Counterfactuals~\cite{ravfogelgumbel} generates counterfactuals by holding sampling noise fixed across completions using the Gumbel-max trick.

\noindent \textbf{Self-Training Paradigm\footnotemark[\value{footnote}]}.
As an alternative to prompt-based methods with large-scale LLMs, we adapt a self-training approach--SRLM~\cite{SRLM}. A smaller LLM--Llama 3.1 8B~\cite{Llama3.1}-- is tuned on its own outputs using Direct Preference Optimization (DPO)~\cite{rafailov2023dpo}, after fine-tuning on the supplementary dataset (Section~\ref{sec:supplementary_set}). 

\begin{table*}[t!]
\centering
\small
\begin{tabularx}{\textwidth}{Xccccc}
\toprule
\textbf{Method} & \textbf{Perplexity $\downarrow$} & \textbf{$\Delta$ Perplexity $\downarrow$} & \textbf{Fwd-Compat. $\uparrow$} & \textbf{Dir. $\uparrow$} & \textbf{FwdCompat-Dir Avg. $\uparrow$} \\
\midrule
Claude 3.5 Haiku & 442.70 & +174.72 & 55.41\% & \underline{73.61\%} & 64.51\% \\
Claude 3.5 Haiku FS & 433.83 & +165.85 & \underline{78.07\%} & 47.54\% & 62.80\% \\
Gemini 2.0 Flash & 459.28 & +191.30 & 53.51\% & 61.99\% & 57.75\% \\
Gemini 2.0 Flash FS & 432.15 & +164.17 & 66.34\% & 58.34\% & 62.34\% \\
Llama4 Maverick & 512.01 & +244.03 & 38.93\% & 50.84\% & 44.89\% \\
Llama4 Maverick FS & 325.36 & +57.38 & 68.92\% & 48.69\% & 58.80\% \\
GPT-4o & 415.27 & +147.29 & 67.84\% & 47.22\% & 57.53\% \\
GPT-4o + FS & 327.03 & +59.05 & \textbf{84.47\%} & 37.76\% & 61.11\% \\
Qwen 2.5 72B & 359.40 & +91.42 & 61.88\% & 57.02\% & 59.45\% \\
Qwen 2.5 72B FS & 302.34 & +34.36 & 73.32\% & 51.27\% & 62.29\% \\
LLMs-for-CFs & 550.25 & +282.28 & 54.75\% & 42.25\% & 48.50\% \\
CounterfactualDistil & 498.31 & +230.33 & 41.23\% & 14.77\% & 28.00\% \\
LM-Counterfactuals & \textbf{155.84} & \textbf{-112.13} & 62.06\% & 68.93\% & \textbf{65.50\%} \\
SRLM & \underline{258.51} & \underline{-9.47} & 56.25\% & \textbf{73.79\%} & \underline{65.02\%} \\
\bottomrule
\end{tabularx}
\caption{Overall evaluation results across all Risk-Opportunity counterfactuals. Best results are bolded, second-best results are underlined. $\downarrow$ (lower score is better), $\uparrow$ (higher score is better). Fwd-Compat. is \textit{forward-compatibility}, Dir. is \textit{directionality}, FwdCompat-Dir Avg. is the average between \textit{forward-comptability} and \textit{directionality} scores. FS stands for few-shot, with results averaged over 5 random samplings for few-shot examples. }
\label{tab:results_summary}
\end{table*}

\section{Results \& Discussion}\label{sec:results}
To inform future model development on this benchmark, we analyze each model's performance and compare them. For clarity, we center our discussion on  $\Delta$ Perplexity and FwdCompat-Dir Avg.--which comprises the average of \textit{forward-compatibility} (Fwd-Compat.) and \textit{directionality} (Dir.) scores (these metrics are defined in Section~\ref{sec:metrics}).

\subsection{Baseline LLM Prompting}

\textbf{LLMs with the highest Fwd-Compat-Dir Avg. scores under zero and few-shot settings do not exhibit the strongest general reasoning performance.}
From Table~\ref{tab:results_summary}, Claude 3.5 Haiku (64.51\%), Claude 3.5 Haiku FS (62.80\%), achieve the highest Fwd-Compat-Dir Avg. scores. This is despite models with lower Fwd-Compat-Dir Avg. performance--Gemini 2.0 Flash, GPT-4o, and Llama 4 Maverick--exhibiting stronger results on general reasoning benchmarks (eg., GPQA)~\cite{Llama-4, yang2024qwen2.5}. This suggests that \textsc{Fin-Force} requires specialized reasoning skills not captured by standard reasoning benchmarks. \textit{Improving \textsc{Fin-Force} performance will require targeting these specialized skills rather than general reasoning alone.}

\begin{figure*}[t!]
    \centering
    \begin{minipage}{0.48\linewidth}
        \centering
        \includegraphics[width=\linewidth]{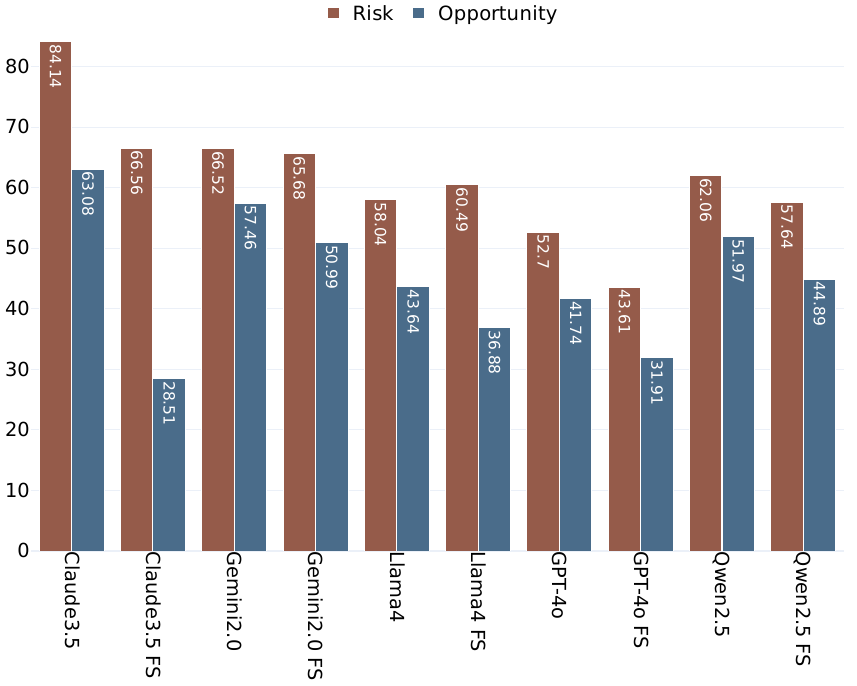}
        \caption{\textit{Directionality} (Dir.) scores between risk and opportunity counterfactuals for different baseline LLM prompting methods under zero and few-shot settings.}
        \label{fig:risk_opp_dir}
    \end{minipage}\hfill
    \begin{minipage}{0.48\linewidth}
        \centering
        \vspace{13.1pt}
        \includegraphics[width=\linewidth]{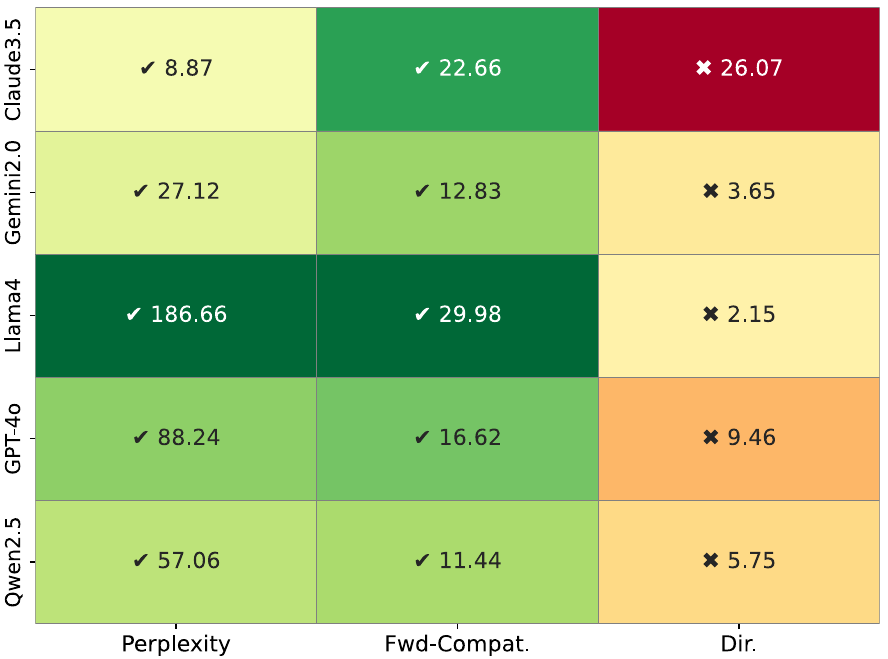}
        \caption{Absolute performance changes with few-shot relative to zero-shot prompting for different LLMs. \ding{51} indicates improvement; \ding{55} indicates degradation.}
        \label{fig:heatmap}
    \end{minipage}
\end{figure*}

\textbf{LLMs underperform at generating directionally accurate opportunity counterfactuals compared to risk counterfactuals, across zero and few-shot settings.} This is shown by the lower \textit{directionality} scores for opportunity compared to risk counterfactuals in Figure~\ref{fig:risk_opp_dir} for all baseline LLM prompting methods. Error analysis ("Trivial Consequences" in Figure~\ref{fig:errors}) reveals that generated opportunity counterfactuals often describe vague or superficial positive market shifts. For example, stating that a firm (Asian Paints) “sees potential in rural markets” without specifying concrete actions that pose an opportunity for market participants. The tendency to default to these superficial shifts suggests that LLMs under zero and few-shot prompting, may lack strong conceptual understanding of meaningful positive market shifts, possibly due to knowledge gaps~\cite{feng2024llmknowledgegap}. \textit{To strengthen this conceptual grasp, advanced prompting such as metacognitive prompting can be explored~\cite{wang2023metacognitive}. This can guide the LLM to explicitly reflect on whether its reasoning exhibits key conceptual principles of a meaningful market shift--i.e. criteria (i) to (iv) of \textit{directionality} in Section~\ref{sec:metrics}--thereby improving performance.} 

\textbf{Few-shot prompting does not improve performance across all metrics.} From Figure~\ref{fig:heatmap}, all LLMs improve on $\Delta$ Perplexity and Fwd-Compat. but show reduced Dir. performance with few-shot compared to zero-shot prompting. This divergence highlights that few-shot prompting cannot optimize performance across all metrics. \textit{Structured prompting strategies--i.e. least-to-most prompting~\cite{zhou2022leasttomost}--may help by decomposing the task into intermediate steps, wherein each metric can be explicitly optimized. This can deliberately guide the model through the reasoning process and reduce the risk of overlooking key criteria.}
 
\subsection{SOTA Counterfactual Generation}\label{sec:sota_cf}

\textbf{SOTA counterfactual prompting strategies underperform significantly, while sampling-based generation performs markedly better.} Despite leveraging specialized prompting strategies for counterfactual text classification and QA, LLMs-for-CF and CounterfactualDistil exhibit among the weakest $\Delta$ Perplexity scores (+282.28 and +230.33) and FwdCompat-Dir Avg. scores (48.50\% and 28.00\%). In contrast, LM-Counterfactuals, which utilizes sampling noise, achieves the best performance ($\Delta$ Perplexity -112.13; FwdCompat-Dir Avg. 65.50\%). \textit{These results suggest that the counterfactual prompting strategies, while effective for other counterfactual tasks that rely on minimal edits or label flipping, do not transfer well to \textsc{Fin-Force}, which demands more substantive reasoning. However, the sampling-based method achieves better generalization for this more complex task.}

\begin{figure*}[t!]
    \centering
    \begin{minipage}[t]{0.48\textwidth}
        \centering
        \includegraphics[width=\linewidth]{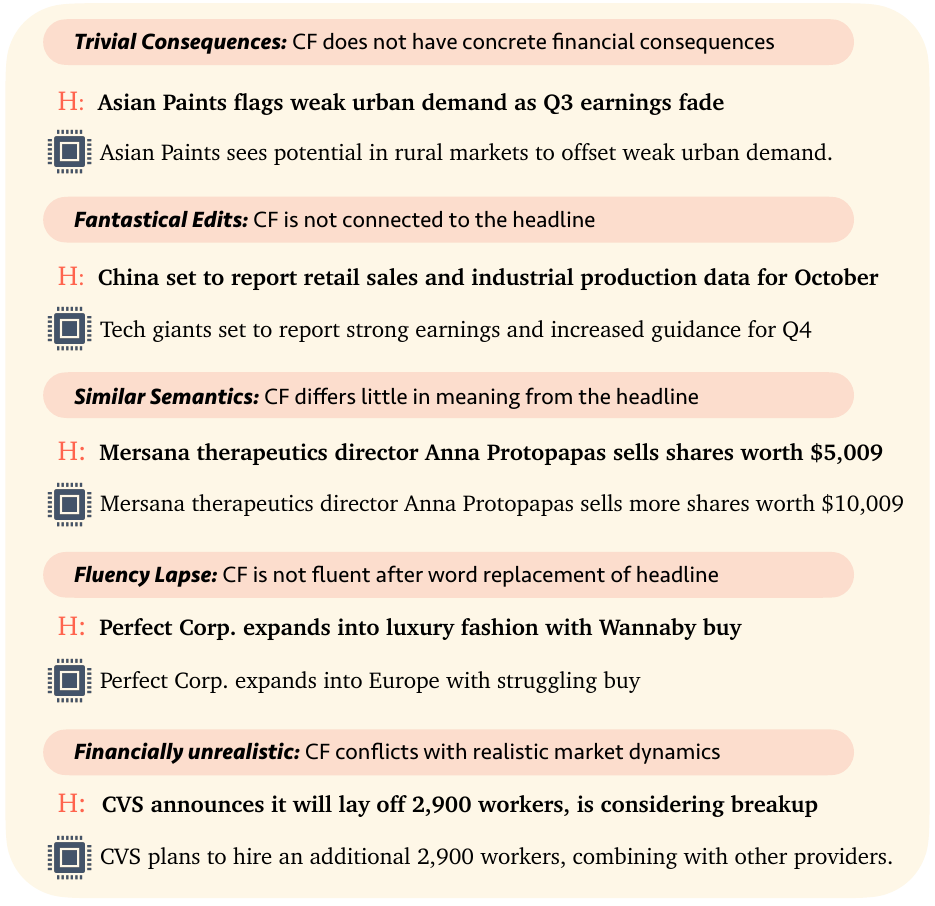}
        \caption{Analysis of prominent error cases. \textcolor{orange}{H} represents a headline in \textsc{Fin-Force}; \raisebox{-0.6ex}{\includegraphics[height=3ex]{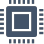}} denotes the errorneous LLM response; CF stands for counterfactual.}
        \label{fig:errors}
    \end{minipage}%
    \hfill
    \begin{minipage}[t]{0.48\textwidth}
        \centering
        \includegraphics[width=\linewidth]{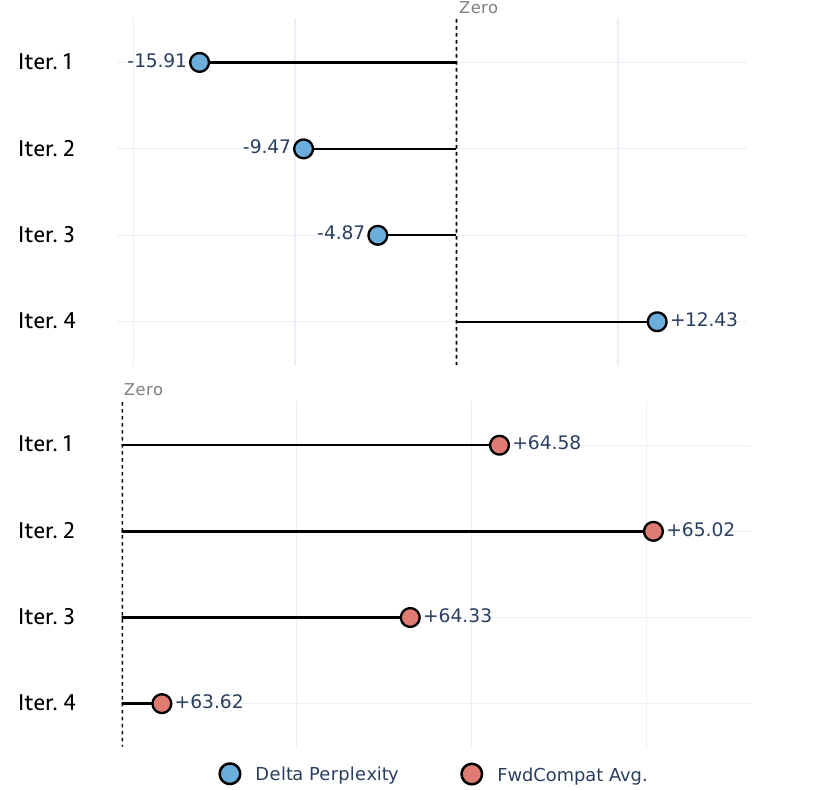}
        \caption{Performance across different iterations of training in the SRLM self-training paradigm. Iter. stands for training iteration.}
        \label{fig:srlm_iter}
    \end{minipage}
\end{figure*}

\textbf{CounterfactualDistil’s masking strategy undermines contextual consistency.} Error analysis ("Fantastical Edits" in Figure~\ref{fig:errors}) shows that masking and replacing key topic words and noun phrases often disrupts the original headline’s narrative. Forced to regenerate content without access to the masked terms, the model often produces counterfactuals that lack contextual fidelity with the original headline, introducing unrelated and disparate market shifts. This compromises both Fwd-Compat. and Dir. scores. \textit{These findings provide insight for future method development on \textsc{Fin-Force} -- preserving the original headline’s context may be important for achieving counterfactuals with high contextual fidelity, which translates to stronger Fwd-Compat. and Dir. performance.}

\textbf{Despite preserving the original headline’s context, LLMs-for-CF produces counterfactuals that lack meaningful positive or adverse market developments, as reflected by its poor Dir. score (third-lowest, at 42.25\%)}. Unlike CounterfactualDistil, LLMs-for-CF reasons over the complete, unmasked headline to infer words to replace, generating counterfactuals that have better contextual consistency with the original headline. However, by restricting changes to word-level substitutions without broader narrative changes (e.g. adding new clauses), the generated counterfactuals often reflect shallow semantic changes from the original headline (i.e. "Similar Semantics" in Figure~\ref{fig:errors}). This restricts the counterfactuals from introducing concrete market shifts that reflect meaningful risk or opportunity, leading to a weaker Dir. score. \textit{These findings suggest that narrative-level rewriting, not just token-level replacement, is required for meaningful and directionally valid counterfactuals.}

\textbf{Word replacement-based methods often compromise the fluency of counterfactual text.} The LLMs-for-CF and CounterfactualDistil methods, which focus on replacing key terms in the original headline, yield among the weakest $\Delta$ Perplexity (+282.28 and +230.33). Error analysis (“Fluency Lapse” in Figure~\ref{fig:errors}) reveals that direct word replacements often produce awkward phrasing or contradictions, as the surrounding sentence is not adapted to accommodate the word changes. This indicates that beyond limiting semantic depth, word replacement strategies also impair fluency and coherence. \textit{Thus, while methods that involve broader narrative rewriting is needed to produce meaningful and directionally valid counterfactuals, they are equally important for ensuring coherence and fluency.}

\textbf{Controlling random sampling noise across counterfactual generations shows promise for enhancing performance.} LM-Counterfactuals controls sampling noise during generation, such that completions primarily reflect prompt changes rather than stochastic variance in the token sampling process. This may help the model follow prompt instructions more consistently (though the exact underlying mechanism requires further study), thereby generating counterfactuals that reflect the best $\Delta$ Perplexity and FwdCompat-Dir Avg scores. Nonetheless, error analysis ("Financially Unrealistic" in Figure~\ref{fig:errors}) shows that the counterfactuals occasionally reflect financially unrealistic reasoning. This affects the validity of their \textit{directionality}, reducing Dir. scores. For example, overly optimistic extensions of negative headlines--i.e. A firm (CVS) planning to hire workers immediately after major layoffs and a potential company breakup. \textit{These findings suggest that controlling sampling noise is a promising direction for counterfactual generation in \textsc{Fin-Force}, given that reducing stochastic variance has been effective for improving performance. However, further gains may depend on enhancing the financial realism of generated counterfactuals. For this, domain adaptation via preference tuning on financial data~\cite{rafailov2023dpo}, can strengthen the financial reasoning of these generations.}

\subsection{Self-Training}\label{sec:results_srlm}

\textbf{Self-training achieves competitive results on \textsc{Fin-Force}.} SRLM achieves the second highest $\Delta$ Perplexity (-9.47) and FwdCompat-Dir Avg. (65.02\%) scores. Notably, SRLM achieves this despite using a relatively simple self-training setup, where the model evaluates its own responses on a basic 10-point scale to construct a preference set for DPO (see Appendix~\ref{appendix:model}). \textit{This strong performance, achieved with a relatively basic design, suggests considerable potential for improvement through more advanced self-training strategies.}

\textbf{Self-training reaches saturation quickly, limiting further improvements in performance.} In SRLM, the model constructs preference sets from its own outputs and trains on them over multiple iterations. From Figure~\ref{fig:srlm_iter}, SRLM reaches its peak FwdCompat Avg. (65.02\%) at the second iteration, and best $\Delta$ Perplexity (-15.91) at the first iteration, with no further gains beyond these points. This effect, known as saturation, is a common limitation of self-training observed across tasks~\cite{wu2024llmmetajudge, SRLM}. However, in \textsc{Fin-Force}, this limitation may be exacerbated by the added complexity of balancing multiple objectives--i.e. \textit{forward-compatibility} (Fwd-Compat.), \textit{directionality} (Dir.), and fluency ($\Delta$ Perplexity)--which makes it harder to reliably distinguish preferred from rejected outputs when building preference sets. \textit{Hence, future development of self-training methods could focus on sustaining effective learning signals and exploring how \textsc{Fin-Force}’s multi-objective complexity--by making it more difficult to judge outputs--affects self-training performance.}

\subsection{Comparison of Methods}
\textbf{Baseline prompting, self-training, and sampling methods offer distinct trade-offs.} Baseline LLM prompting--using LLMs in zero and few-shot settings--delivers competitive performance without fine-tuning and can be deployed in plug-and-play fashion (i.e. directly through APIs without GPU training). In contrast, tuning-based methods (SRLM) and sampling-based algorithms (LM-Counterfactuals) can achieve better performance on \textsc{Fin-Force} but involve greater computational complexity (i.e. training pipelines, decoding control) for preference tuning and controlled decoding. \textit{Therefore, each method entails practical trade-offs that extend beyond performance optimization alone, and should be considered in relation to computational cost, complexity, and resource availability.}

\textbf{From Table~\ref{tab:results_summary}, SRLM surpasses the performance of all baseline LLM prompting methods on $\Delta$ Perplexity (-9.47) and FwdCompat-Dir Avg. (65.02\%).} This highlights how a self-training method (SRLM) can enable smaller LLMs (Llama 3.1-8B) to achieve strong \textsc{Fin-Force} performance, mitigating the need to rely on larger models (i.e. GPT-4o, Llama4-Maverick). With performance no longer a limiting factor, this enables the adoption of smaller models that offer greater controllability and are more computationally efficient at inference time--qualities that are often impractical with larger LLMs. \textit{Therefore, not only does SRLM achieve strong performance, it also enables the development of efficient, controllable smaller language models tailored to \textsc{Fin-Force}, reducing dependence on large-scale LLMs.}

\textbf{The sampling-based LM-Counterfactuals and SRLM methods are not mutually exclusive, and can potentially be integrated to optimize performance.} In fact, combining them can address a key limitation of LM-Counterfactuals noted previously in Section~\ref{sec:sota_cf}--generating counterfactuals that occasionally reflect unrealistic financial reasoning. Self-training can help mitigate this limitation by adapting the LM-Counterfactuals model to the financial domain. As part of self-training, an LLM-judge can automatically construct a preference set from LM-Counterfactuals’ outputs, selecting responses with sound financial reasoning and rejecting those without. The model can then be optimized on this preference set through DPO, reducing financial reasoning lapses and improving counterfactual generation quality. \textit{Therefore, integrating LM-Counterfactuals with self-training (SRLM) may thus represent a promising direction for improving model performance on \textsc{Fin-Force}.}

\section{Conclusion}
We introduced \textsc{Fin-Force}, a novel benchmark for forward counterfactual generation in finance, and evaluated a range of methods to offer insights for future model development. Through experiments, we find that while existing methods offer a starting point, they face limitations in producing fluent, forward-compatible, and directionally valid counterfactuals. Of these methods, sampling-based generation achieved the highest overall performance, self-training enabled smaller models to attain competitive performance, while zero and few-shot prompting for selected LLMs offered a plug-and-play approach with strong results. However, all methods have limitations--errors and inconsistent performance on metrics--highlighting the need for further work. Our work aims to lay the foundation for scalable, automated insights into potential market opportunities and risks for stakeholders.

\section*{Limitations}
Our benchmark is currently limited to English-language headlines. While this provides broad coverage, we acknowledge that multilingual financial news is important for global market analysis--particularly in regions where English sources are limited. Future work will explore extending the benchmark to include non-English news sources. Additionally, while the benchmark focuses on delivering counterfactual insights for financial stakeholders, counterfactuals have also been applied to improve the explainability and performance of AI models~\cite{treviso2023crest,mothilal2020explainingcounterfactuals}, which is an ongoing concern in critical sectors such as finance and healthcare~\cite{mengaldo2024explain, turbe2025tell}. To this end, future work may also explore leveraging forward counterfactuals to improve the explainability and performance of financial models (i.e. for stock prediction).

\section*{Ethical Considerations}
Data collection procedures received approval from our research group’s internal ethics review board. We uphold ethical standards by collecting and processing data with strict attention to privacy and confidentiality. Our data sources include publicly available online news obtained via NewsAPI. As these news items may reference companies or individuals, we take care to anonymize sensitive or personal information in the \textsc{Fin-Force} benchmark, focusing exclusively on the financial content relevant to our study. The models employed are also publicly accessible and sourced from published research. We comply with the copyright terms set by the respective holders for all data, software packages, and models used. Human annotators operate under rigorous guidelines to ensure objectivity and minimize bias. We are committed to transparency in our methodology and clearly attribute all sources to uphold ethical standards in data usage and sharing. 
Dataset and code are released publicly for research purposes only, in accordance with relevant copyright requirements.

\section*{Acknowledgements}
This research/project is supported by the NUS Sustainable and Green Finance Institute (SGFIN), NUS Asian Institute of Digital Finance (AIDF), Ministry of Education, Singapore under its MOE Academic Research Fund Tier 2 (STEM RIE2025 Award MOE-T2EP20123-0005: “Neurosymbolic AI for Commonsense-based Question Answering in Multiple Domains”), MOE Tier 2 Award (MOE-T2EP50221-0006: “Prediction-to-Mitigation with Digital Twins of the Earth’s Weather”), MOE Tier 1 Award (MOE-T2EP50221-0028: “Discipline-Informed Neural Networks for Interpretable Time-Series Discovery”), and by the RIE2025 Industry Alignment Fund – Industry Collaboration Projects (IAF-ICP) (Award I2301E0026: “Generative AI"), administered by A*STAR, as well as supported by Alibaba Group and NTU Singapore

\bibliography{main}

\appendix

\section{Benchmark}\label{appendix:benchmark}

\subsection{Keywords for Querying Market News}
As described in Section~\ref{sec:news_collection}, we collect a broad corpus of news data by querying NewsAPI using relevant keywords. The retrieved headlines are then annotated to select those included in our benchmark. We specify the full list of keywords in Table~\ref{tab:market_keywords}.

\begin{table}[h!]
\centering
\begin{tabular}{@{}p{\linewidth}@{}}
\toprule
\textbf{Market Keywords} \\
\midrule
GDP growth \\
Interest rates \\
Inflation rate \\
Unemployment rate \\
Consumer spending \\
Disposable income \\
Stock market performance \\
Foreign exchange rates \\
Economic growth forecasts \\
Investment climate \\
Cost of capital \\
Economic stability \\
\bottomrule
\end{tabular}
\caption{Full list of market-related keywords used to query NewsAPI for collecting financial news headlines.}\label{tab:market_keywords}
\end{table}

\subsection{Financial Market Categories}
Our benchmark comprises 1368 news headlines that belong to defined financial market categories. Each of these categories describes a specific type of financial market event. We provide a full description of these categories in Table~\ref{tab:market_categories_description}.

\subsection{Benchmark Samples}
To provide a better appreciation of the financial news headlines collected in the \textsc{Fin-Force} benchmark, we provide more samples in Table~\ref{tab:benchmark_samples}. Additionally, we denote the financial market category that each sample belongs to.



\subsection{Supplementary Dataset}\label{appendix:supplementary_dataset}
To generate the forward counterfactuals in the supplementary dataset, we generate a risk and opportunity counterfactual from a single news headline utilizing GPT-4o. We leverage the same implementation of GPT-4o under zero-shot setting as detailed in Appendix~\ref{appendix:model} (baseline LLM prompting methods). 

\subsection{Annotators}
A total of five human contributors, all based in Singapore and recruited from reputable research institutions, participated in the benchmark annotation (Section~\ref{sec:annotation_process}) and human-LLM agreement study (Section~\ref{sec:metrics}) as annotators or verifiers. They are affiliated with the Asian Institute of Digital Finance and the National University of Singapore’s College of Design and Engineering. All annotators are actively pursuing doctoral or post-doctoral research in finance and possess expertise in financial markets. Their participation in the annotation process forms part of their formal academic and research activities. They were compensated at a rate exceeding the local minimum wage (SGD \$15/hour). The annotators adhered strictly to the established annotation scheme and guidelines and provided consent for the dataset’s use in research. To ensure objectivity in the human-LLM agreement study for \textit{directionality} and \textit{forward-compatibility} metrics, the annotators were deliberately kept independent from the paper's development.

\subsection{News Annotation Instructions}
In this section, we present the full annotation instructions provided to annotators for labeling the news headlines in our benchmark. The annotators are first instructed to read the background, followed by the general instructions, event status and material relevance guidelines. \\

\noindent \textbf{Background:} Financial news headlines describe a wide range of information about the economy and financial markets. These headlines can provide actionable intelligence for market stakeholders, helping to inform decision-making. However, the content of financial headlines varies greatly, from opinions and commentary to concrete events and factual developments. In this annotation task, you will analyze financial news headlines and evaluate them according to specific criteria. Your annotations will be used for research purposes.\\

\noindent \textbf{General Instructions:}

\begin{enumerate}
    \item Please annotate each headline according to whether it describes a financial market event.
    \item In order for a headline to qualify as describing a financial market event, it must meet two criteria: \textit{Event Status} and \textit{Material Relevance}. 
    \item A headline satisfies the \textit{Event Status} criteria if it satisfies all the requirements outlined in the \textit{Event Status Guidelines}. A headline satisfies the \textit{Material Relevance} criteria if it satisfies all the requirements outlined in the \textit{Material Relevance Guidelines}.
    \item First, evaluate whether the headline satisfies all the requirements in \textit{Event Status Guidelines}.
    \item If the headline satisfies all the requirements in \textit{Event Status Guidelines}, assess whether the headline satisfies all the requirements in the \textit{Material Relevance Guidelines}.
    \item Only if the headline meets both criteria--\textit{Event Status} and \textit{Material Relevance}--classify the headline as "TRUE". This denotes that the headline describes a financial market event.
    \item If the headline meets neither criteria or only meets one criteria, mark it as "FALSE". This denotes that the headline does not describe a financial market event.
    \item For headlines where you are not confident about the annotation, please flag them. These will be discussed and further reviewed by the annotation team. \\
\end{enumerate} 

\noindent \textbf{Event Status Guidelines:}

\begin{enumerate}
    \item The headline must explicitly describe an ongoing financial market event. This is defined as a current, factual development representing a specific change or occurrence affecting financial markets or the broader economy.
    \item Examples of financial market events include but are not limited to central bank decisions, earnings reports releases, regulatory changes, geopolitical escalations, or macroeconomic data releases.
    \item Exclude headlines that provide only generic commentary or opinions without describing a concrete development.
\end{enumerate} 

\noindent \textbf{Material Relevance Guidelines:}

\begin{enumerate}
    \item The headline must describe an financial market event that is material or meaningful to financial market participants, indicating that it can potentially influence financial decision-making or financial market outcomes.
    \item Material or meaningful financial market events include, but are not limited to, broad systemic drivers (e.g., monetary policy changes, macroeconomic indicators) and significant company-specific events (e.g., mergers, earnings surprises, corporate restructuring). To guide this assessment, we provide a detailed list of financial market categories that encompass these types of events: \textit{Monetary Policy \& Central Banking, Corporate Strategy \& Operations, Geopolitical \& Regulatory Developments, Financial Markets \& Asset Performance, Supply Chain \& Logistics, ESG \& Sustainability Developments, Technology \& Innovation, Labour \& Employment, Macroeconomic Indicators, Banking \& Financial Stability.} [Note: We provided the annotators with Table~\ref{tab:market_categories_description}, which provides definitions of these categories.]
    \item The financial market event highlighted in the headline must explicitly relate to at least one of the aforementioned categories to be considered material.
    \item If the event explicitly relates to at least one of the categories, please mark the category that the headline is most closely associated with.
\end{enumerate}

\subsection{Human-LLM Agreement Study}
Besides the news annotation, we also conducted a human-LLM agreement study. The goal was to evaluate whether the LLM judge (GPT-4o) aligns with the preferences of proficient human annotators when using the criteria of \textit{forward-compatibility} and \textit{directionality} to evaluate counterfactuals. The human-LLM agreement study did not exactly follow the same News Annotation Process (described in Section~\ref{sec:annotation_process}). Specifically, we retained only the (1) Training Phase, and the annotations were conducted over a three-day period. We did not implement disagreement resolution or verification procedures, as the goal was to measure average alignment across the annotators and the LLM, rather than to produce a fully adjudicated ground truth. 

In the following, we present the full annotation instructions provided to the annotators for the study. Similar to the news annotation, the annotators are first instructed to read the background, followed by the general instructions, \textit{forward-compatibility} and \textit{directionality} guidelines. A total of 500 counterfactual samples, randomly selected from all experiments in Section~\ref{sec:expts}, were evaluated. Each counterfactual is paired with its original financial news headline, which is also taken into account during evaluation, as we will detail below.\\ 

\noindent \textbf{Background:} In this annotation task, you will evaluate counterfactuals generated by LLMs according to specific criteria. Each counterfactual must be assessed with consideration of the original financial news headline from which it was generated. Your annotations will be used for research purposes. \\

\noindent \textbf{General Instructions:} 
\begin{enumerate}
    \item For each counterfactual, evaluate it in relation to its original news headline, as required by the \textit{Forward-Compatibility} and \textit{Directionality} criteria.
    \item Assess the counterfactual using the two evaluation criteria: \textit{Forward-Compatibility} and \textit{Directionality}.
    \item First, determine whether the counterfactual satisfies all the requirements specified in the \textit{Forward-Compatibility Guidelines}. If it does, mark Forward-Compatibility as "TRUE"; otherwise, mark it as "FALSE".
    \item Then, assess whether the counterfactual satisfies the requirements in the \textit{Directionality Guidelines}. If it does, mark Directionality as "TRUE"; otherwise, mark it as "FALSE".
    \item Record the outcome for each criterion--\textit{Forward-Compatibility} and \textit{Directionality}--separately. The counterfactual should receive a TRUE or FALSE label for each criterion independently of the other.
\end{enumerate} 

\noindent \textbf{Forward-Compatibility Guidelines:}
\begin{enumerate}
\item The counterfactual must represent a plausible future development that logically follows from the market event described in the original news headline. This entails that the counterfactual meets the requirements of \textit{consistent progression} and \textit{non-contradiction}.
\item The counterfactual must have \textit{consistent progression}. The counterfactual must reflect a continuation logically connected to the original market event. For example, if the original headline reports an interest rate hike, a counterfactual about unrelated environmental regulations would not qualify.
\item The counterfactual must have \textit{non-contradiction}. The counterfactual must not negate or reverse the original market event by introducing mutually exclusive scenarios. For example, if the original headline states that a company is exiting a market, a counterfactual suggesting the company is expanding in that same market would be contradictory.
\end{enumerate}

\noindent \textbf{Directionality Guidelines:}

\begin{enumerate}
\item The counterfactual must reflect a clear and meaningful market shift--either toward improvement (opportunity) or deterioration (risk) in market conditions--relative to the market event described in the original news headline. This entails that the counterfactual meets the requirements of \textit{relative significance}, \textit{logical soundness}, \textit{scope of impact} and \textit{financial consequence}.
\item The counterfactual must have \textit{relative significance.} The counterfactual must reflect a shift that is substantial compared to the original market event. Superficial changes--such as minor wording differences or trivial updates that do not alter the market implications--do not qualify.
\item The counterfactual must have \textit{logical soundness.} The counterfactual must reflect a shift that is grounded in logical reasoning. It cannot reflect implausibility in financial or economic logic. 
\item The counterfactual must have sufficient \textit{scope of impact}. The counterfactual must reflect a shift that affects not just isolated parties, but also other market participants. For example, a company initiative to upgrade office equipment would not qualify, while a CEO change that could affect investors, competitors, or market expectations would.
\item The counterfactual must have \textit{financial consequence}. The counterfactual must reflect a shift that entails financial effects likely to influence market behavior or decision-making. Examples include but are not limited to changes affecting revenue, costs, investment, or market valuation.
\end{enumerate}

\section{Experiments}\label{appendix:expts}

\subsection{LLM-as-a-judge Evaluation}
Two of the evaluation metrics for \textsc{Fin-Force}, \textit{directionality} and \textit{forward-compatibility} leverage GPT-4o classification under an LLM-as-a-judge framework~\cite{zheng2023judgingllm}. Each generated counterfactual is presented to the LLM one at a time for evaluation, and separate runs are performed for assessing \textit{forward-compatibility} and \textit{directionality}. We highlight the LLM-as-a-judge prompts utilized for \textit{directionality} in Table~\ref{tab:llm_judge_directionality}, and for \textit{forward-compatibility} in Table~\ref{tab:llm_judge_fc}.

\subsection{Model Implementation Details}\label{appendix:model}
We will detail the implementation of the experimental methods in Section~\ref{sec:expts}. For the baseline LLM prompting methods, we leverage the respective LLM versions: gpt-4o-2024-08-06 for GPT4o~\cite{gpt-4o}, claude-3-5-haiku-20241022 for Claude 3.5 Haiku~\cite{claude-3.5-haiku}, gemini-2.0-flash for Gemini 2.0 Flash~\cite{gemini-2.0-flash}, Llama 4 Maverick (17Bx128E) for Llama 4 Maverick~\cite{Llama-4}, Qwen 2.5-72B-Instruct for Qwen 2.5 72B~\cite{yang2024qwen2.5}. For these methods, we use the prompt template in Table~\ref{tab:prompt_template}, and include few-shot examples for few-shot setups while omitting them for zero-shot setups. We provide samples of our few shot examples in Table~\ref{tab:few_shot_examples}. To ensure consistency, few-shot examples are kept the same across the different LLMs. 

For the SOTA counterfactual generation methods, we adapt LLMs-for-CFs~\cite{nguyen-etal-2024-llms_for_cfs}, CounterfactualDistil~\cite{feng-etal-2024-counterfactualdistllation}, LM-Counterfactuals~\cite{ravfogelgumbel} to our task. The original LLMs-for-CF implementation uses chain-of-thought prompting to identify and replace keywords for counterfactual generation, and includes a one-shot example. In our setup, we adopt a similar approach--guiding the model to identify and replace keywords--while also providing a reasoning chain example. We implement this using GPT-4o as the underlying LLM. The prompt template for our implementation of LLMs-for-CF is shown in Table~\ref{tab:prompt_template}. 

The original CounterfactualDistil implementation comprises of two main steps. First, it masks the topic word and noun phrases in the original text. Then, it prompts an LLM to generate replacements for the masked spans based on a predefined target label, ultimately producing the counterfactual. In our setup, we use GPT-4o to identify topic words using the prompt in Table~\ref{tab:cfdistil_topicprompt}, and apply the SpaCy library~\cite{honnibal2020spacy} to extract noun phrases. These text spans are then masked. The masked text is then passed to GPT-4o to generate replacements according to the target risk and opportunity labels. The full prompt template for our implementation of CounterfactualDistil at the final counterfactual generation step is shown in Table~\ref{tab:prompt_template}. 

The original LM-Counterfactuals method generates counterfactuals by fixing the sampling noise across both the base and counterfactual generations, using the Gumbel-max trick. It first generates a base completion to recover the sampling noise, which is then held fixed when generating the counterfactual. In our setup, this base completion is formulated as a continuation of the original financial news headline, using the prompt in Table~\ref{tab:gumbel_base_prompt}. The recovered noise is then reused to generate both risk and opportunity counterfactuals. For this counterfactual generation step, we utilize a similar task prompt to the one used in the baseline LLM prompting methods, as shown in Table~\ref{tab:prompt_template}, without few-shot examples. Following the original implementation, we use the Llama-3-8B-Instruct model from Hugging Face\footnote{https://huggingface.co/meta-llama/Meta-Llama-3-8B-Instruct}, applying 4-bit quantization and keeping the same hyperparameters.

For self-training, we adapt the SRLM framework~\cite{SRLM}. In the original implementation, an LLM is first fine-tuned on human-annotated alignment data, then used to generate synthetic prompts and responses. The LLM also judges these responses to create a synthetic preference set, which is used to further train the model via Direct Preference Optimization (DPO)~\cite{rafailov2023dpo}. This process is repeated over multiple iterations. In our setup, we train a 4-bit quantized Llama3.1-8B-Instruct model from Hugging Face\footnote{https://huggingface.co/unsloth/Meta-Llama-3.1-8B-Instruct-bnb-4bit} using LoRA and the Unsloth library. After fine-tuning on our supplementary dataset (Section~\ref{sec:supplementary_set}), the model is prompted with the prompt shown in Table~\ref{tab:srlm_news_prompt} to generate synthetic financial headlines. It is then prompted with a similar task prompt to the one used in the baseline LLM prompting methods (Table~\ref{tab:prompt_template}), without few-shot examples, to produce risk and opportunity counterfactuals. These counterfactual outputs are judged by the LLM using a point-scale prompt (Table~\ref{tab:srlm_judge_prompt}), similar to the original implementation, to distinguish chosen and rejected outputs for constructing the DPO preference set. The rest of our implementation follows the original SRLM with two modifications to the hyperparameters. First, due to the lack of multi-source validation data, we use validation loss for early stopping (patience = 2 steps), due to its simplicity and established role in model selection. Second, to accommodate computational constraints, we train the model with LoRA using lora\_dropout=0.1, lora\_alpha=16, and lora\_r=64, and approximate the original effective batch size of 16 by using a batch size of 1 with 16 gradient accumulation steps. Additionally, following the SRLM setup, we train the model for two iterations and report results from the second iteration in Table~\ref{tab:results_summary}. We continue training the model for a third and fourth iteration to examine the saturation effect, as discussed in Section~\ref{sec:results_srlm}.

\subsection{Computation and Tools Used for Our Study}
This study was conducted with the help of external, publicly available tools (NewsAPI\footnote{https://newsapi.org}, Pytorch\footnote{https://pytorch.org}, Huggingface\footnote{https://huggingface.co}, SpaCy~\cite{honnibal2020spacy}, GPT-4o, Llama 3, Llama 4, Claude 3.5 Haiku, Qwen 2.5, Gemini 2.0 Flash), with all experiments run on a single NVIDIA GeForce RTX 4090 GPU. To compute perplexity, we use GPT-2 from the evaluate\footnote{https://huggingface.co/docs/evaluate/en/index} package. GPT-4o is deployed through the OpenAI API\footnote{https://platform.openai.com/docs/overview}, while Claude 3.5 Haiku, Llama 4 Maverick, Qwen 2.5 72B and Gemini 2.0 Flash are deployed through the Openrouter API\footnote{https://openrouter.ai}. We specify the model sizes utilized for our study (model sizes for proprietary models are unavailable): Qwen 2.5 72B (72B), GPT-2 (1.5B), Llama 4 Maverick (400B), Llama3.1-8B-Instruct (8B), Llama3-8B-Instruct (8B).

\subsection{Results Breakdown}
Since the \textsc{Fin-Force} task requires generating both a risk and an opportunity counterfactual for each news headline, the overall results in Table~\ref{tab:results_summary} report the average evaluation scores across all generated risk and opportunity counterfactuals. For a more detailed performance breakdown, Table~\ref{tab:risk_results_summary} and Table~\ref{tab:opportunity_results_summary} report the evaluation scores for all risk and opportunity counterfactuals, respectively.

\begin{table*}[ht]
\centering
\begin{tabularx}{\textwidth}{|l|X|}
\hline
\textbf{Category} & \textbf{Description} \\
\hline
Monetary policy \& central banking & Involves central bank actions, interest rate decisions, and policy announcements that systematically influence market liquidity and investor sentiment. \\
\hline
Corporate strategy \& operations & Pertains to strategic business decisions, restructuring, and operational changes that impact corporate performance and market dynamics. \\
\hline
Geopolitical \& regulatory developments & Covers political events, regulatory changes, or international developments that systematically affect market stability and economic policy. \\
\hline
Financial markets \& asset performance & Relates to trends, shifts, or events in financial markets that directly drive asset valuations and investment portfolios. \\
\hline
Supply chain \& logistics & Encompasses disruptions or improvements in global supply chains and logistics that systematically influence corporate earnings and market risk. \\
\hline
ESG \& sustainability developments & Focuses on environmental, social, and governance initiatives that are shaping long-term market trends and systematic risk assessments. \\
\hline
Technology \& innovation & Addresses technological breakthroughs and innovations that are driving structural shifts in competitiveness and economic growth. \\
\hline
Labour \& employment & Deals with changes in employment trends and labour market dynamics that have systematic implications for consumer spending and economic performance. \\
\hline
Macroeconomic indicators & Involves key economic statistics and trends that provide systematic insights into overall market conditions. \\
\hline
Banking \& financial stability & Concerns developments within the banking sector or financial system that influence systemic risk and market confidence. \\
\hline
\end{tabularx}
\caption{Descriptions of key market categories in the \textsc{Fin-Force} benchmark.}
\label{tab:market_categories_description}
\end{table*}

\begin{table*}[ht]
\centering
\begin{tabularx}{\textwidth}{@{}X l@{}}
\toprule
\textbf{News Headline} & \textbf{Market Category} \\
\midrule
Google, Microsoft Are Spending Massively on AI, Quarterly Earnings Show. & Technology \& innovation \\
\midrule
UK's largest retailers warn Budget will lead to job cuts. & Labour \& employment \\
\midrule
Japan approves new climate, energy and industry policies through 2040. & ESG \& sustainability developments \\
\midrule
U.S. tariffs on steel, aluminum spark strong backlash across Europe. & Geopolitical \& regulatory developments \\
\midrule
MAS eases monetary policy for the second time this year; lowers core inflation forecast. & Monetary policy \& central banking \\
\bottomrule
\end{tabularx}
\caption{Sample financial news headlines and their associated market categories.}
\label{tab:benchmark_samples}
\end{table*}

\begin{table*}[t]
\small
\begin{tabularx}{\textwidth}{p{2cm}X}
\toprule
\textbf{Evaluation} & \textbf{Prompt} \\
\midrule
Directionality\newline(Opportunity) & 
You are an expert judge assessing whether a counterfactual generated by an LLM represents a materially favorable market development relative to the original market development described in the news headline.\newline
The news headline is enclosed between <news> and </news>.\newline
The counterfactual is enclosed between <response> and </response>.
\newline \newline
Evaluation Criterion:\newline
Your task is to determine whether the counterfactual reflects a clear and meaningful improvement in market conditions for market participants relative to the original market development. Return true only if all the following criteria are met:\newline
- The counterfactual must clearly describe a market-relevant improvement with a positive impact on market participants more broadly, beyond any isolated party involved.\newline
- The market-relevant improvement must be clearly tied to financial consequences. General positivity or reputational gain (e.g., “increased trust”) is not sufficient unless explicitly linked to tangible market benefits.\newline
- The improvement must be material relative to the original market development, representing a clear positive shift relative to the original market development, not a superficial or marginal gain.\newline
- The counterfactual must be logically coherent and causally sound. Improvements based on implausible, inconsistent, or unsupported reasoning should be rejected.\newline
- The improvement must be clearly stated and well-supported. Vague, implied, or unsubstantiated benefits do not qualify.\newline
- The counterfactual must introduce or amplify a credible opportunity for market participants--including but not limited to enhanced revenue, improved efficiency, capital access, or favorable positioning--that could reasonably influence decision-making or market behavior.\newline \newline
Return your evaluation in the following JSON format:\newline
\{ \newline
\hspace*{1em}"directionality": \{ \newline
\hspace*{2em}"value": <true or false> \newline
\hspace*{1em}\} \newline
\} \\
\midrule
Directionality\newline(Risk) & 
You are an expert judge assessing whether a counterfactual generated by an LLM represents a materially adverse market development relative to the original market development described in the news headline.\newline
The news headline is enclosed between <news> and </news>.\newline
The counterfactual is enclosed between <response> and </response>.
\newline \newline
Evaluation Criterion\newline
Your task is to determine whether the counterfactual reflects a clear and meaningful deterioration in market conditions for market participants relative to the original market development. Return true only if all the following criteria are met:\newline
- The counterfactual must clearly describe a market-relevant deterioration with a negative impact on market participants more broadly, beyond any isolated party involved.\newline
- The market-relevant deterioration must be clearly tied to financial consequences. General negativity or reputational harm (e.g., “loss of trust”) is not sufficient unless explicitly linked to tangible financial consequences.\newline
- The deterioration must be material relative to the original market development. It should reflect a clear negative shift relative to the original market development, not a superficial or minor setback.\newline
- The counterfactual must be logically coherent and causally sound. Deteriorations based on implausible, inconsistent, or unsupported reasoning should be rejected.\newline
- The deterioration must be clearly stated and well-supported. Vague, implied, or unsubstantiated harms do not qualify.\newline
- The counterfactual must introduce or amplify a credible risk for market participants--including but not limited to uncertainty, volatility, or exposure to future loss--that could reasonably affect decision-making or market behavior.\newline \newline
Return your evaluation in the following JSON format:\newline
\{ \newline
\hspace*{1em}"directionality": \{ \newline
\hspace*{2em}"value": <true or false> \newline
\hspace*{1em}\} \newline
\} \\
\bottomrule
\end{tabularx}
\caption{LLM-judge prompts for evaluating the \textit{directionality} of counterfactuals. Directionality (Opportunity) is used to evaluate opportunity counterfactuals, while Directionality (Risk) is used to evaluate risk counterfactuals.}
\label{tab:llm_judge_directionality}
\end{table*}

\begin{table*}[t]
\small
\vspace{-0.5cm}
\begin{tabularx}{\textwidth}{X}
\toprule
\textbf{Prompt} \\
\midrule
You are an expert judge tasked with assessing the quality of the counterfactual response generated by an LLM, based on provided financial news headlines.\newline
- The news headline is enclosed between <news> and </news>.\newline
- The LLM response (the counterfactual) is enclosed between <response> and </response>.\newline
\newline
Your task is to carefully review the LLM-generated counterfactual and assess whether it represents a plausible future development that remains consistent with the original news headline. You should return a structured evaluation using the criterion below.\newline
\newline
Evaluation Criterion:\newline
Forward Compatibility, or in other words, does the counterfactual represent a plausible future development from the original market development described in the news headline? Respond with a true or false value. To be considered forward-compatible, the counterfactual must satisfy the following criteria:\newline
- The counterfactual must describe a development that could plausibly take place after the original news event.\newline
- The counterfactual must not cancel out the original market development. This includes but is not limited to introducing mutually exclusive outcomes or implying that the original market development did not happen. \newline
\newline
Return your evaluation in the following JSON format:\newline
\{\newline
\hspace*{1em}"forward\_compatibility": \{\newline
\hspace*{2em}"value": <true or false>\newline
\hspace*{1em}\}\newline
\} \\
\bottomrule
\end{tabularx}
\caption{LLM-judge prompt for evaluating the \textit{forward-compatibility} of counterfactuals.}
\label{tab:llm_judge_fc}
\end{table*}

\begin{table*}[t]
\small
\begin{tabularx}{\textwidth}{p{3.5cm}X}
\toprule
\textbf{Method(s)} & \textbf{Prompt} \\
\midrule
Claude 3.5 Haiku, Gemini 2.0 Flash, GPT-4o, LLaMA 4 Maverick, Qwen 2.5 72B, LM-Counterfactuals, Self-Training (SRLM) & 
You are a financial expert tasked with generating minimally edited counterfactuals based on a provided financial headline that describes a market development.\newline
Your goal is to generate a risk counterfactual and an opportunity counterfactual, following from our requirements below:\newline
\newline
Risk Counterfactual:\newline
- Minimally edit the original market development headline to represent a plausible alternate counterfactual that represents an adverse shift in the market development.\newline
- The adverse shift should reflect an adverse market outcome or deterioration in market conditions.\newline
- The alternate counterfactual must be forward-looking, which means that it can plausibly occur after the original market development.\newline
\newline
Opportunity Counterfactual:\newline
- Minimally edit the original market development headline to represent a plausible alternate counterfactual that represents a positive shift in the market development.\newline
- The positive market shift reflects a beneficial market outcome or improvement in market conditions.\newline
- The alternate counterfactual must be forward-looking, which means that it can plausibly occur after the original market development.\newline
\newline
Input format:\newline
- A single financial news headline.\newline
\newline
Your output must be valid JSON matching this structure. Do not make explicit numeric predictions or quantitative outcomes.\newline
\{\newline
\hspace*{1em}"Counterfactuals": [\newline
\hspace*{2em}\{\newline
\hspace*{3em}"original\_headline": "Article Headline",\newline
\hspace*{3em}"opportunity\_counterfactual": "Opportunity Counterfactual",\newline
\hspace*{3em}"risk\_counterfactual": "Risk Counterfactual"\newline
\hspace*{2em}\}\newline
\hspace*{1em}]\newline
\}  \newline 
\newline \{\textbf{Few-Shot Examples}\}
\newline \newline Input: \{\textbf{Financial News Headline}\} \\
\bottomrule
\end{tabularx}
\caption{Prompt template used by each method for counterfactual generation -- continued on the next page}
\label{tab:prompt_template}
\end{table*}

\begin{table*}[t]
\ContinuedFloat
\small
\begin{tabularx}{\textwidth}{p{3.5cm}X}
\toprule
\textbf{Method(s)} & \textbf{Prompt} \\
\midrule
LLMs-for-CF & 
You are a financial expert tasked with generating minimally edited counterfactuals based on a provided financial headline that describes a market development.\newline
Your goal is to generate a risk counterfactual and an opportunity counterfactual, following from our requirements below:\newline
\newline
Risk Counterfactual:\newline
- Minimally edit the original market development headline to represent a plausible alternate counterfactual that represents an adverse shift in the market development.\newline
- The adverse shift should reflect an adverse market outcome or deterioration in market conditions.\newline
- The alternate counterfactual must be forward-looking, which means that it can plausibly occur after the original market development.\newline
\newline
Opportunity Counterfactual:\newline
- Minimally edit the original market development headline to represent a plausible alternate counterfactual that represents a positive shift in the market development.\newline
- The positive market shift reflects a beneficial market outcome or improvement in market conditions.\newline
- The alternate counterfactual must be forward-looking, which means that it can plausibly occur after the original market development.\newline
\newline
Please follow these reasoning steps before returning your output:\newline
1. Identify key phrases or words that signal the core market development in the original headline.\newline
2a. Change these key phrases or words to construct a forward-looking adverse market counterfactual with minimal changes.\newline
2b. Change these key phrases or words to construct a forward-looking positive market counterfactual with minimal changes.\newline
3a. Replace the key phrases and words from step 1 in the original text by the key phrases and words in step 2a, returning a risk counterfactual that reflects an adverse market shift.\newline
3b. Replace the key phrases and words from step 1 in the original text by the key phrases and words in step 2b, returning an opportunity counterfactual that reflects a positive market shift.\newline
\newline
Example Reasoning Chain:\newline
(Original headline)\newline
"Apple announces expansion of iPhone production in India"\newline
\newline
Step 1. Identify key phrases:\newline
- "announces expansion"\newline
- "iPhone production"\newline
- "in India"\newline
\newline
Step 2a. Edits for risk:\newline
- "announces expansion" → "faces delay in expansion"\newline
\newline
Step 2b. Edits for opportunity:\newline
- "announces expansion" → "accelerates expansion"\newline
\newline
Step 3a:\newline
"Apple faces delay in expansion of iPhone production in India"\newline
Step 3b:\newline
"Apple accelerates expansion of iPhone production in India"\newline
\newline
Final output format:\newline
\{\newline
\hspace*{1em}"original\_headline": "Article Headline",\newline
\hspace*{1em}"reasoning": "Reasoning Chain",\newline
\hspace*{1em}"opportunity\_counterfactual": "Opportunity Counterfactual",\newline
\hspace*{1em}"risk\_counterfactual": "Risk Counterfactual"\newline
\} \\
\bottomrule
\end{tabularx}
\caption{Prompt template used by each method for counterfactual generation -- continued on the next page}
\end{table*}

\begin{table*}[t]
\ContinuedFloat
\small
\begin{tabularx}{\textwidth}{p{3.5cm}X}
\toprule
\textbf{Method(s)} & \textbf{Prompt} \\
\midrule
CounterfactualDistil & 
You are a financial expert tasked with generating minimally edited counterfactuals based on a provided financial headline that describes a market development.\newline
Your goal is to generate a risk counterfactual and an opportunity counterfactual, following from our requirements below:\newline
\newline
Risk Counterfactual:\newline
- Minimally edit the original market development headline to represent a plausible alternate counterfactual that represents an adverse shift in the market development.\newline
- The adverse shift should reflect an adverse market outcome or deterioration in market conditions.\newline
- The alternate counterfactual must be forward-looking.\newline
\newline
Opportunity Counterfactual:\newline
- Minimally edit the original market development headline to represent a plausible alternate counterfactual that represents a positive shift in the market development.\newline
- The positive market shift reflects a beneficial market outcome or improvement in market conditions.\newline
- The alternate counterfactual must be forward-looking.\newline
\newline
You will be given a masked version of the original headline, with placeholders like [MASK] in key slots (e.g., actors, sectors, verbs, or descriptors).\newline
These masked tokens are meant to be minimally edited while reflecting a directional shift (risk or opportunity).\newline
Please complete the [MASK] part of the headlines based on the specified opportunity and risk direction, to make it a counterfactual with smooth semantics and clear logic.\newline
\newline
Example:\newline
Input: "[MASK] markets set to [MASK] ahead of [MASK] and [MASK] decisions this week, [MASK] and [MASK] markets [MASK]"\newline
Output:\newline
\{\newline
\hspace*{1em}"original\_masked\_headline": "[MASK] markets set to [MASK] ahead of [MASK] and [MASK] decisions this week, [MASK] and [MASK] markets [MASK]",\newline
\hspace*{1em}"opportunity\_counterfactual": "Japan markets set to rally ahead of Fed and BOJ decisions this week, Australia and China markets reopen",\newline
\hspace*{1em}"risk\_counterfactual": "Japan markets set to plunge ahead of Fed and BOJ decisions this week, Australia and China markets remain closed"\newline
\} \\
\bottomrule
\end{tabularx}
\caption{Prompt template used by each method for counterfactual generation.}
\end{table*}

\begin{table*}[t]
\small
\begin{tabularx}{\textwidth}{X}
\toprule
\textbf{Few-shot examples} \\
\midrule
Input: Canada unveils multibillion-dollar plan to cut carbon emissions.\newline
\newline
Output:\newline
\hspace*{2em}original\_headline: Canada unveils multibillion-dollar plan to cut carbon emissions.\newline
\hspace*{2em}risk\_counterfactual\_scenario: Canada's multibillion-dollar emissions reduction plan faces setbacks due to regulatory challenges.\newline
\hspace*{2em}opportunity\_counterfactual\_scenario: Spurred by initial multibillion-dollar plan's success, Canada further invests in renewable energy to boost economic growth and reduce carbon emissions. \\
\midrule
Input: Germany's bond yields set for biggest monthly jump in over a decade - Reuters.\newline
\newline
Output:\newline
\hspace*{2em}original\_headline: Germany's bond yields set for biggest monthly jump in over a decade - Reuters.\newline
\hspace*{2em}risk\_counterfactual\_scenario: Germany's bond yields rise more rapidly than expected, raising concerns of potential economic downturn - Reuters.\newline
\hspace*{2em}opportunity\_counterfactual\_scenario: Germany's bond yields stabilize after recent jump, signaling improved investor confidence - Reuters. \\
\midrule
Input: Hartree Partners invests in nature-based voluntary carbon offset projects - Reuters.\newline
\newline
Output:\newline
\hspace*{2em}original\_headline: Hartree Partners invests in nature-based voluntary carbon offset projects - Reuters.\newline
\hspace*{2em}risk\_counterfactual\_scenario: Hartree Partners' investment in nature-based voluntary carbon offset projects faces regulatory hurdles.\newline
\hspace*{2em}opportunity\_counterfactual\_scenario: Hartree Partners' investment in nature-based voluntary carbon offset projects yields significant environmental returns. \\
\midrule
Input: Nikkei rides high while traders wait on US inflation.\newline
\newline
Output:\newline
\hspace*{2em}original\_headline: Nikkei rides high while traders wait on US inflation.\newline
\hspace*{2em}risk\_counterfactual\_scenario: Nikkei falters as US inflation data stokes market fears.\newline
\hspace*{2em}opportunity\_counterfactual\_scenario: Nikkei soars further as US inflation shows signs of cooling. \\
\midrule
Input: US government debt reaches new milestone.\newline
\newline
Output:\newline
\hspace*{2em}original\_headline: US government debt reaches new milestone.\newline
\hspace*{2em}risk\_counterfactual\_scenario: US government debt milestone triggers investor concerns over economic stability.\newline
\hspace*{2em}opportunity\_counterfactual\_scenario: US government debt milestone prompts strategic fiscal policy reforms. \\
\bottomrule
\end{tabularx}
\caption{Few-shot examples used for counterfactual generation.}
\label{tab:few_shot_examples}
\end{table*}

\clearpage

\begin{table*}[t]
\small
\begin{tabularx}{\textwidth}{X}
\toprule
\textbf{Prompt} \\
\midrule
You are a topic word extractor. Your task is to extract the most relevant topic word from the given text.\newline
\newline
Input: \{Financial News Headline\} \\
\bottomrule
\end{tabularx}
\caption{Prompt for inferring topic words from the news headline, as part of the CounterfactDistil method.}
\label{tab:cfdistil_topicprompt}
\end{table*}

\begin{table*}[t]
\small
\begin{tabularx}{\textwidth}{X}
\toprule
\textbf{Prompt} \\
\midrule
News: \{Financial News Headline\}.\newline \newline
You are a financial expert tasked with reasoning about plausible future developments based on this headline.\newline
Generate a minimally edited, forward-looking continuation that remains coherent with the original market development. \\
\bottomrule
\end{tabularx}
\caption{Prompt for generating the base continuation from the news headline, as part of the LM-Counterfactuals method.}
\label{tab:gumbel_base_prompt}
\end{table*}

\begin{table*}[t]
\small
\begin{tabularx}{\textwidth}{X}
\toprule
\textbf{Prompt} \\
\midrule
<task> Come up with one new financial news headline. Write only the financial news headline, with no further text or explanation.\newline \newline
The examples below are enclosed in <example></example> tags. </task> \\
\bottomrule
\end{tabularx}
\caption{Prompt for generating synthetic news headlines, as part of the SRLM method.}
\label{tab:srlm_news_prompt}
\end{table*}

\begin{table*}[t]
\small
\begin{tabularx}{\textwidth}{X}
\toprule
\textbf{Prompt} \\
\midrule
Review the base news and the corresponding counterfactual scenarios using the additive 10-point scoring system described below.\newline
\newline
The original news is enclosed between <news> and </news>, and the generated scenario is enclosed between <response> and </response>.\newline
\newline
Points are accumulated based on the satisfaction of each criterion:\newline
- Add 1 point if the risk counterfactual is topically relevant and reflects a modification or extrapolation of the original news.\newline
- Add another point if the opportunity counterfactual is also relevant to the original news.\newline
- Add 1 point if the risk counterfactual makes minimal edits to the original news (preserving structure and intent).\newline
- Add another point if the opportunity counterfactual also uses minimal edits appropriately.\newline
- Add 1 point if the risk counterfactual is forward-compatible, describing a plausible future development that does not contradict the original news.\newline
- Add another point if the opportunity counterfactual is also forward-compatible.\newline
- Add 1 point if the risk counterfactual clearly reflects an adverse market development.\newline
- Add another point if the opportunity counterfactual clearly reflects a favorable market development.\newline
- Add 1 point if the risk counterfactual is cohesive, meaning it is clear, logical, and internally consistent.\newline
- Add another point if the opportunity counterfactual is cohesive in the same way.\newline
- If either counterfactual is incoherent, irrelevant, or fails to fulfill any criteria, award 0 points.\newline
\newline
<news>\{news\}</news>\newline
<response>\{response\}</response>\newline
\newline
After examining the news and the counterfactual scenario:\newline
- output the score of the evaluation using this exact format: "score: <total points>", where <total points> is between 0 and 10\newline
- Briefly justify your total score, up to 100 words. \\
\bottomrule
\end{tabularx}
\caption{Prompt for self-judging counterfactual responses to create a synthetic preference set, as part of the SRLM method.}
\label{tab:srlm_judge_prompt}
\end{table*}

\clearpage

\begin{table*}[h!]
\centering
\small
\begin{tabularx}{\textwidth}{Xccccc}
\toprule
\textbf{Method} & \textbf{Perplexity $\downarrow$} & \textbf{$\Delta$ Perplexity $\downarrow$} & \textbf{Fwd-Compat. $\uparrow$} & \textbf{Dir. $\uparrow$} & \textbf{FwdCompat-Dir Avg. $\uparrow$} \\
\midrule
Claude 3.5 Haiku & 448.56 & +180.58 & 37.94\% & \textbf{84.14\%} & 61.04\% \\
Claude 3.5 Haiku FS & 432.73 & +164.75 & 65.90\% & 66.56\% & 66.23\% \\
Gemini 2.0 Flash & 434.38 & +166.40 & 53.44\% & 66.52\% & 59.98\% \\
Gemini 2.0 Flash FS & 396.48 & +128.51 & 67.07\% & 65.68\% & \underline{66.38\%} \\
Llama4 Maverick & 504.64 & +236.67 & 32.31\% & 58.04\% & 45.18\% \\
Llama4 Maverick FS & 341.86 & +73.88 & 65.72\% & 60.49\% & 63.10\% \\
GPT-4o & 389.76 & +121.78 & 68.57\% & 52.70\% & 60.64\% \\
GPT-4o FS & 327.14 & +59.16 & \textbf{84.84\%} & 43.61\% & 64.22\% \\
Qwen 2.5 72B & 352.13 & +84.15 & 63.08\% & 62.06\% & 62.57\% \\
Qwen 2.5 72B FS & 294.73 & +26.75 & \underline{74.78\%} & 57.64\% & 66.21\% \\
LLMs-for-CFs & 514.41 & +246.43 & 47.81\% & 47.59\% & 47.70\% \\
CounterfactualDistil & 527.83 & +259.85 & 38.96\% & 21.71\% & 30.34\% \\
LM-Counterfactuals & \textbf{193.58} & \textbf{-74.39} & 57.46\% & 77.56\% & \textbf{67.51\%} \\
SRLM & \underline{253.92} & \underline{-14.06} & 44.30\% & \underline{78.87\%} & 61.59\% \\
\bottomrule
\end{tabularx}
\caption{Evaluation results across all risk counterfactuals. Best results are bolded, second-best results are underlined. $\downarrow$ (lower score is better), $\uparrow$ (higher score is better). Fwd-Compat. is \textit{forward-compatibility}, Dir. is \textit{directionality}, FwdCompat-Dir Avg. is the average between \textit{forward-comptability} and \textit{directionality} scores. FS stands for few-shot, with results averaged over 5 random samplings for few-shot examples.}
\label{tab:risk_results_summary}
\end{table*}

\begin{table*}[h!]
\centering
\small
\begin{tabularx}{\textwidth}{Xccccc}
\toprule
\textbf{Method} & \textbf{Perplexity $\downarrow$} & \textbf{$\Delta$ Perplexity $\downarrow$} & \textbf{Fwd-Compat. $\uparrow$} & \textbf{Dir. $\uparrow$} & \textbf{FwdCompat-Dir Avg. $\uparrow$} \\
\midrule
Claude 3.5 Haiku & 436.83 & +168.86 & 72.88\% & \underline{63.08\%} & \underline{67.98\%} \\
Claude 3.5 Haiku FS & 434.93 & +166.95 & \textbf{90.24\%} & 28.51\% & 59.37\% \\
Gemini 2.0 Flash & 484.18 & +216.20 & 53.58\% & 57.46\% & 55.52\% \\
Gemini 2.0 Flash FS & 467.82 & +199.84 & 65.61\% & 50.99\% & 58.30\% \\
Llama4 Maverick & 519.38 & +251.40 & 45.54\% & 43.64\% & 44.59\% \\
Llama4 Maverick FS & 308.85 & +40.87 & 72.11\% & 36.88\% & 54.50\% \\
GPT-4o & 440.77 & +172.79 & 67.11\% & 41.74\% & 54.43\% \\
GPT-4o FS & 326.92 & +58.94 & \underline{84.10\%} & 31.91\% & 58.01\% \\
Qwen 2.5 72B & 366.67 & +98.69 & 60.67\% & 51.97\% & 56.32\% \\
Qwen 2.5 72B FS & 309.94 & +41.96 & 71.86\% & 44.89\% & 58.37\% \\
LLMs-for-CFs & 586.10 & +318.13 & 61.70\% & 36.92\% & 49.31\% \\
CounterfactualDistil & 468.78 & +200.81 & 43.49\% & 7.82\% & 25.66\% \\
LM-Counterfactuals & \textbf{118.10} & \textbf{-149.88} & 66.67\% & 60.31\% & 63.49\% \\
SRLM & \underline{263.12} & \underline{-4.86} & 68.20\% & \textbf{68.71\%} & \textbf{68.46\%} \\
\bottomrule
\end{tabularx}
\caption{Evaluation results across all opportunity counterfactuals. Best results are bolded, second-best results are underlined. $\downarrow$ (lower score is better), $\uparrow$ (higher score is better). Fwd-Compat. is \textit{forward-compatibility}, Dir. is \textit{directionality}, FwdCompat-Dir Avg. is the average between \textit{forward-comptability} and \textit{directionality} scores. FS stands for few-shot, with results averaged over 5 random samplings for few-shot examples.}
\label{tab:opportunity_results_summary}
\end{table*}

\end{document}